\documentclass[conference]{IEEEtran}
\usepackage{amsmath}
\usepackage{bm}
\usepackage{amsthm}
\usepackage{amsfonts}
\usepackage{algorithm}
\usepackage{algorithmic}
\usepackage{graphicx}
\usepackage{multirow}
\usepackage{subfigure}
\usepackage[normalem]{ulem}
\usepackage{color}
\usepackage{url}

\useunder{\uline}{\ul}{}

\usepackage{array}
\newcolumntype{L}[1]{>{\raggedright\let\newline\\\arraybackslash\hspace{0pt}}m{#1}}
\newcolumntype{C}[1]{>{\centering\let\newline  \\\arraybackslash\hspace{0pt}}m{#1}}
\newcolumntype{R}[1]{>{\raggedleft\let\newline \\\arraybackslash\hspace{0pt}}m{#1}}

\graphicspath{ {img/} }

\theoremstyle{definition}

\def\etal{{\em et al.\;}}

\begin{document}

\title{XOGAN: One-to-Many Unsupervised Image-to-Image Translation}

\author{\IEEEauthorblockN{Yongqi Zhang}
\IEEEauthorblockA{Department of Computer Science and Engineering \\
Hong Kong University of Science and Technology \\
Clear Water Bay, Hong Kong\\
		\{yzhangee\}@cse.ust.hk}
}

\maketitle

\begin{abstract}
Unsupervised image-to-image translation aims at learning the relationship between samples from two image domains without supervised pair
information. The relationship between two domain images can be one-to-one, one-to-many or many-to-many. In this paper, we study the one-to-many
unsupervised image translation problem in which an input sample from one domain can correspond to multiple samples in the other domain. To learn the complex relationship between the two domains, we introduce an additional variable to control the variations in our one-to-many mapping. A
generative model
with an \textit{XO}-structure, called the XOGAN, is proposed to learn the cross domain relationship among the two domains and the additional
variables. Not only can we learn to translate between the two image domains, we can also handle the translated images with additional variations.
Experiments are performed on unpaired image generation tasks, including edges-to-objects translation and facial image translation. We show that
the proposed XOGAN model can generate plausible images and control variations, such as color and texture, of the generated images. Moreover, 
while state-of-the-art unpaired image generation algorithms tend to generate images with monotonous colors,
XOGAN
can generate more diverse results.
\end{abstract}

\vspace{5px}


\begin{IEEEkeywords}
Unsupervised image translation, Image generation, Generative adversarial networks, One-to-many mapping
\end{IEEEkeywords}

\section{Introduction}
\label{sec-intro}

With the development of deep generative models, image generation has become popular in various applications. Among them, image-to-image translation, which learns a mapping from one domain to another, is a recent hot topic. Many computer vision tasks, including cross-domain image generation \cite{isola2016image, kim2017learning, taigman2016unsupervised, zhu2017unpaired, yi2017dualgan}, super-resolution \cite{ledig2017photo}, colorization \cite{yao2015color,zhang2016colorful}, image inpainting \cite{yeh2016semantic}, image style transfer \cite{gatys2016image, li2017universal, johnson2016perceptual}, can be considered as image translation. Based on the two domains of images (either paired or unpaired), a generative model is trained to learn their relationship. 

\begin{figure}[htbp]
\centering
\includegraphics[width=0.45\textwidth]{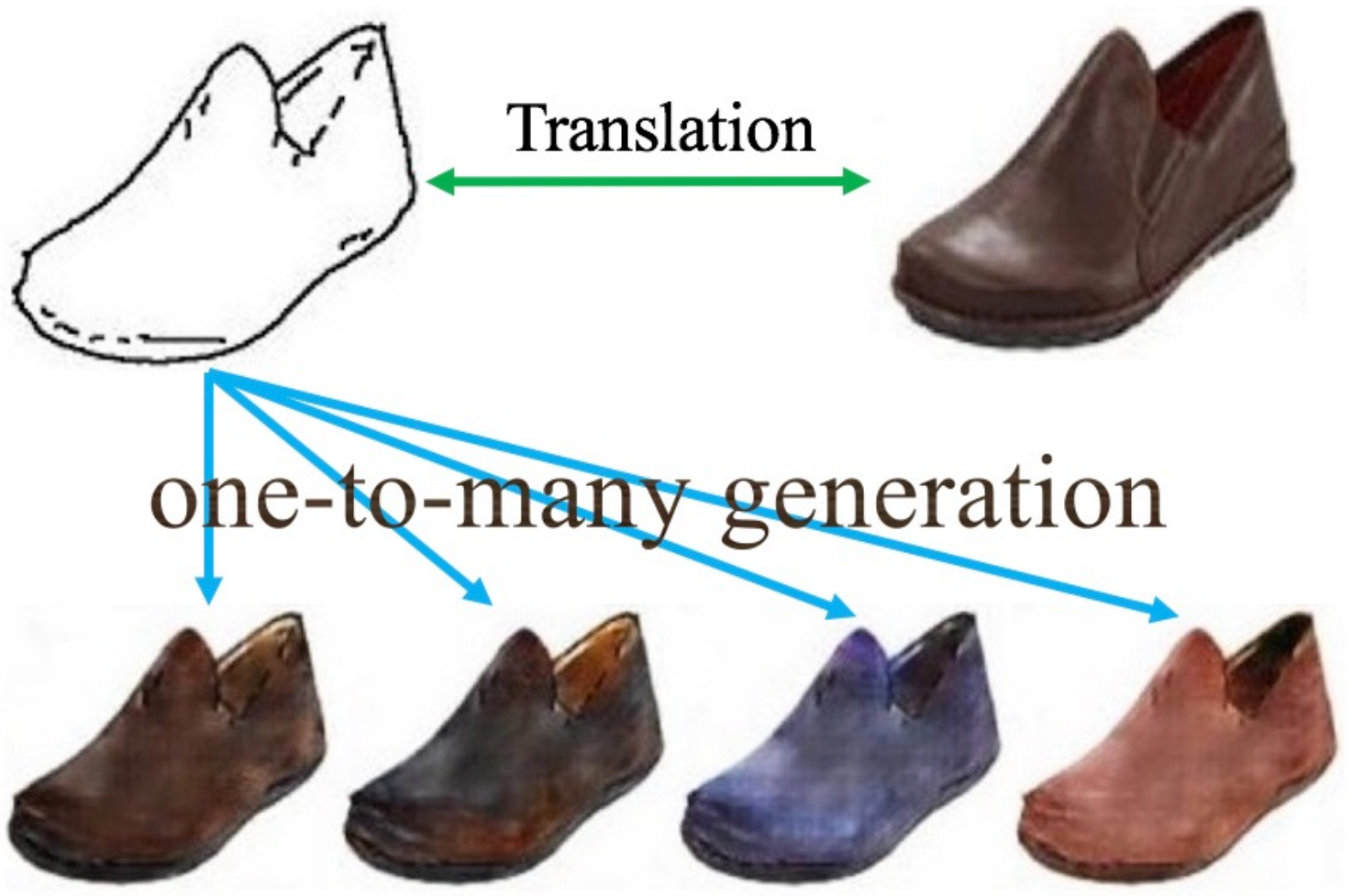}
\caption{Example one-to-many translation. Here, the translation is between a set of edge images and a set of real shoes. The proposed model can learn the one-to-many generation of multiple colored shoes from one edge input.}
\label{fig-example}
\end{figure}

Image translation can be defined as follows. Given an image $X_S$, we map it to a target domain image  $X_T$ that shares some similarity or has
close relationship with $X_S$. The task is to learn the mapping $f: X_S\rightarrow X_T$ that transforms the source domain distribution $p(X_S)$
to the target domain distribution $p(X_T)$. For example, $X_T$ can be a scaled-up version of $X_S$ in super-resolution, or $X_T$ is a colored
version of a grayscale image $X_S$, or $X_T$ is a photo of its sketch $X_S$ (Figure \ref{fig-example}). In these tasks, we assume that there also
exists an inverse mapping $g: X_T \rightarrow X_S$. With the help of the inverse mapping, we can understand better about the two domains by
learning their joint relationship. 

Given a paired data set, early approaches learn the mapping by using the input-output pairs in a supervised manner \cite{isola2016image, ledig2017photo}. The main challenge is to learn and generalize the relationship between the given pairs. However, paired data sets are usually hard to collect, and the corresponding target domain image may not even exist in practice. For example, it is hard to connect the set of paired images between photos and artistic works. Another example is that if the two domains are male faces and female faces, then there does not exist paired data of a same person. In these cases, supervised models fail because of the lack of a ground truth mapping for training. Learning unpaired image translation is more practical and has received more attention recently \cite{kim2017learning, taigman2016unsupervised, zhu2017unpaired, yi2017dualgan}. Moreover, unsupervised (or unpaired) image generation is more practical since data collection is much easier. 


In this paper, we consider the task of unsupervised one-to-many image generation. Given two image domains $\mathcal A$ and $\mathcal B$, we learn
to transfer images in domain $\mathcal A$ to domain $\mathcal B$, and vice versa. Different from previous works on unpaired image translation \cite{kim2017learning, taigman2016unsupervised, zhu2017unpaired}, we assume that there can be many possible target images in domain $\mathcal B$ given the same image in domain $\mathcal A$. For example, in the edge-to-shoe translation task in Figure 1, there can be different colors and textures when generating shoes. To model this variation, we propose to use an additional variable $Z$ to complement images in domain $\mathcal A$. Moreover, this variable $Z$ can be easily sampled from a prior distribution, such as the normal distribution. 

To learn the relationship among $\mathcal A, \mathcal B$ and $Z$,
we propose a novel generative model under the constraint of domain adversarial loss and cycle consistency loss, which is first defined in \cite{zhu2017unpaired}.
The proposed model, which
will be called XOGAN,
is assembled in an ``XO''-structure, and is trained under the generative adversarial network (GAN) framework \cite{goodfellow2014generative, gulrajani2017improved, mirza2014conditional, salimans2016improved, arjovsky2017wasserstein}.


In our experiments, we show results on generating shoes and handbags with diverse colors and textures given the edge images. Besides, when the
additional variable is kept the same for different edge inputs, we can generate objects with the same colors. Moreover, we can alter the colors
of different objects by substituting the variable $Z$. Hence,
not only 
can our model generate plausible images as in other generative models, it can also replace the color of a certain image with another. 

Section \ref{sec-related} first reviews the related works. The proposed ``XO''-structure and the training procedure are introduced in Section \ref{sec-model}. Section \ref{sec-exp} presents experimental results on the {\sf edges2shoes}, {\sf edges2handbags} and {\sf CelebA} data sets.
Finally, Section \ref{sec-conclude} gives some concluding remarks and future directions.


\vspace{.1in}
\noindent
{\em Notations}.
Samples from the two image domains are denoted by $A\in \mathbb R_{\mathcal A}^{h\times w\times c}$ and $B \in \mathbb R_{\mathcal B}^{h\times
w\times c}$, where the subscripts $\mathcal A$ and $\mathcal B$ are the domain indicators,
and $h, w, c$ 
are the height, width, and channel,
respectively. The empirical distributions of the image domains are denoted $P_{\mathcal A}$, and $P_{\mathcal B}$, respectively. The additional
variable $Z\in \mathbb R_{Z}^d$ is sampled from a standard normal distribution $P_{Z}=\mathcal N(0, I)$, where $d$ is the dimensionality of $Z$. 

The generator for image domain $\mathcal A$ is denoted $G_{A}: \mathbb R_{\mathcal B}^{h\times w\times c}\rightarrow \mathbb R_{\mathcal A}^{h\times w\times
c}$, in which the subscripts denote the image domains. The generator for image domain $\mathcal B$ is denoted $G_{B}: (\mathbb R_{\mathcal A}^{h\times w\times c}, \mathbb R_{Z}^{d})\rightarrow \mathbb R_{\mathcal B}^{h\times w\times c}$. The input is a concatenation of the image from domain
$\mathcal A$ and  a sample from the distribution $P_{Z}$. The generator (also called an encoder) for $Z$ is denoted $G_Z: \mathbb R_{\mathcal B}^{h\times w\times c}\rightarrow \mathbb R_{Z}^{d}$. 

The first path generated fake samples in domain $\mathcal A$ and $\mathcal B$ are denoted $\bar{A}$ and $\bar{B}$, and the encoded variable is $\bar{Z}$. When 
the fake samples $\bar{A}, \bar{B}, \bar{Z}$ 
are forwarded
once again through the generators, 
the output variables in each domain 
are denoted
$\hat{A}, \hat{B}, \hat{Z}$ (Figure \ref{fig-model-gen}).

The discriminator networks are denoted $D_A: \mathbb R_{\mathcal A}^{h\times w\times c}\rightarrow [0, 1]$, $D_B: \mathbb R_{\mathcal B}^{h\times
w\times c}\rightarrow [0, 1]$ and $D_Z: \mathbb R_{Z}^{d}\rightarrow [0, 1]$, respectively.

\section{Related Work}
\label{sec-related}

\subsection{Generative Adversarial Networks}

The generative adversarial network (GAN) \cite{goodfellow2014generative} is a powerful generative model that can generate plausible images. The
GAN contains two modules: a generator $G$ that generates samples and a discriminator $D$ that tries to distinguish whether the sample is from the
real or generated distribution. The generator aims to confuse the discriminator by generating samples that are difficult to differentiate from
the real ones. Training of GAN often suffers from issues such as vanishing gradient and mode collapse \cite{salimans2016improved}, in which the
generator tends to collapse to points in a single mode. Very recently, a number of techniques have been introduced to stabilize training
procedure \cite{gulrajani2017improved, salimans2016improved}. In cross-domain image generation \cite{isola2016image, kim2017learning,
zhu2017unpaired, liu2017unsupervised}, GAN is a powerful tool to match the generated image to the real image distributions, especially when
paired images are not available. 


\subsection{Supervised Image-to-Image Translation}

Isola \etal \cite{isola2016image} showed that cross-domain image translation can be learned and generalized  using a paired data set. By using the conditional GAN \cite{mirza2014conditional}, their model can generate plausible photographs from sketches or semantic layouts. Zhu \etal \cite{zhu2017toward} 
used bicycle consistency between the latent code and output images to generate multi-modal target domain images.  However, paired data sets are
not always available, and unpaired data sets are more common in practice.

\subsection{Unsupervised Image-to-Image Translation}
\label{sec-unpaired}

Taigman \etal \cite{taigman2016unsupervised} introduced the domain transfer network (DTN) to generate emoji-style images from facial images in an
unsupervised manner. In the DTN, image translation is a one-way mapping. If we train another model to map the emoji images back to real faces,
the face identity may be changed. More recently, bidirectional mapping becomes more appealing, and has been studied in the DiscoGAN
\cite{kim2017learning}, CycleGAN \cite{zhu2017unpaired} and DualGAN \cite{yi2017dualgan}. These models use one generator and one discriminator
for each mapping, and the symmetric structure helps to learn the bidirectional mapping. 

\begin{figure}[htbp]
	\centering
	\subfigure[Forward path. \label{fig:forward}] {\includegraphics[width=0.27\textwidth]{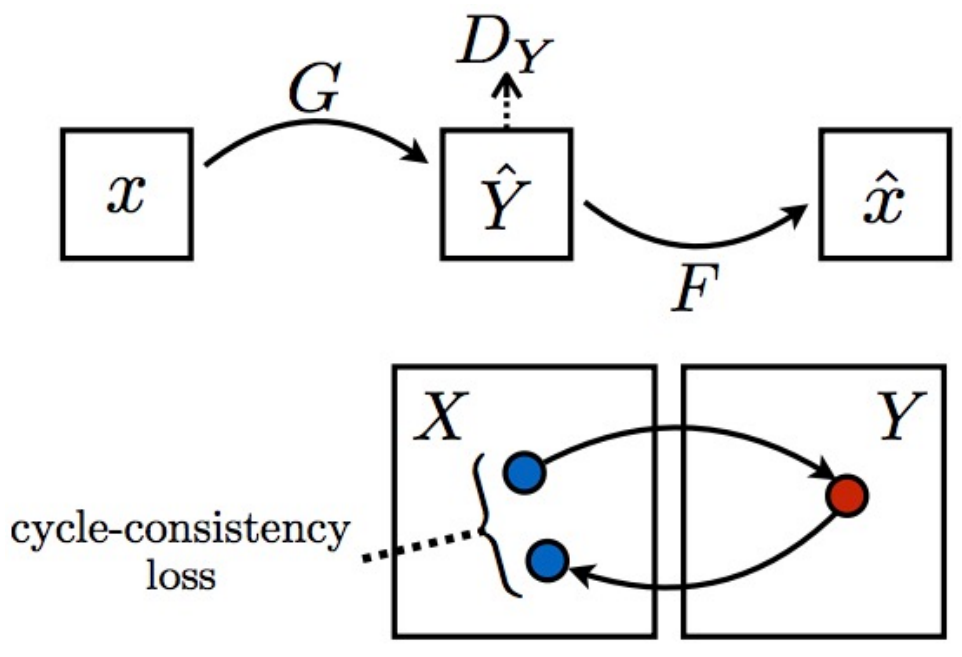}}
	
	\subfigure[Backward path. \label{fig:back}]
	{\includegraphics[width=0.26\textwidth]{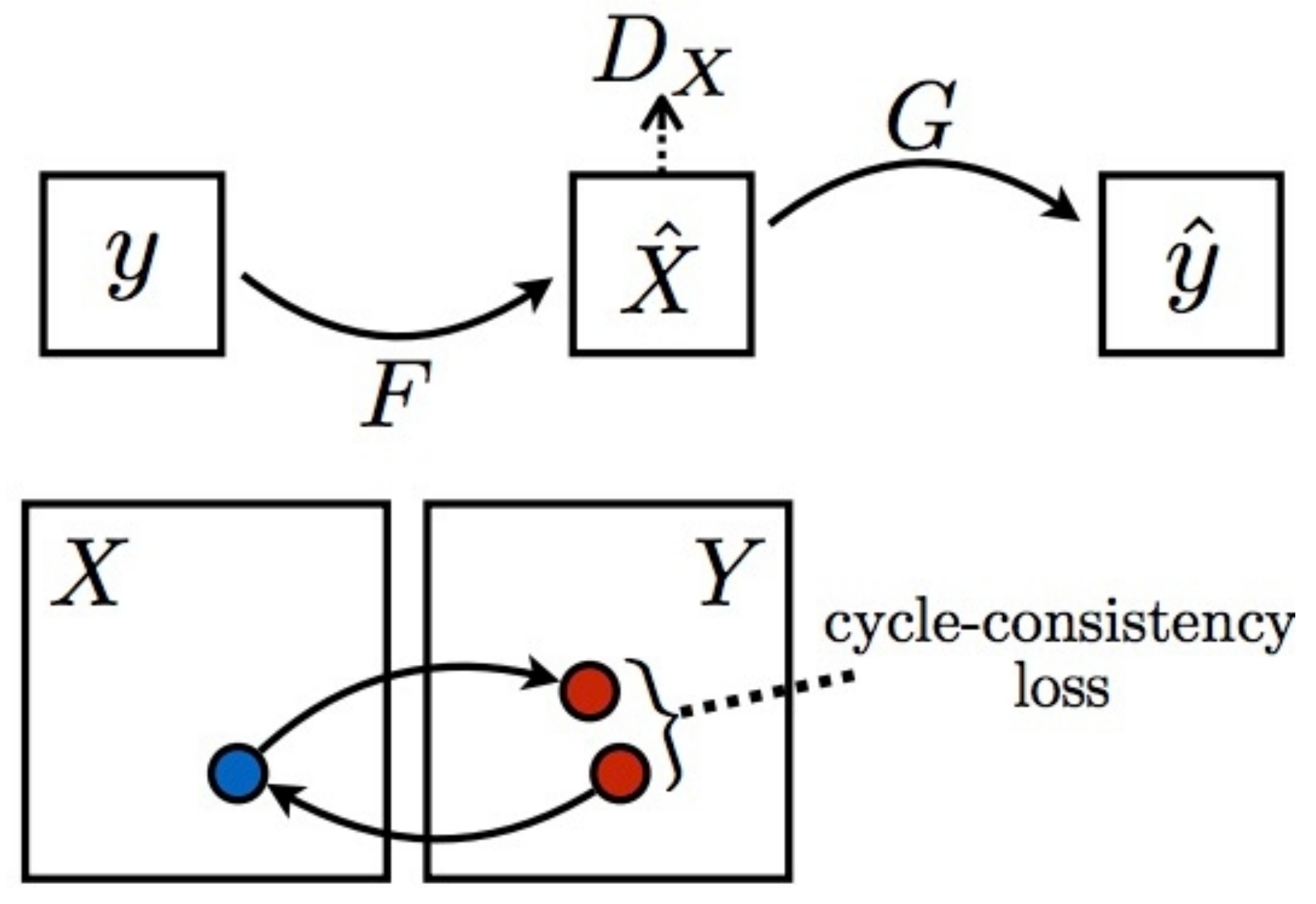}}
	\caption{The CycleGAN \cite{zhu2017unpaired}.}
	\label{fig-model-cycle}
\end{figure}

Figure~\ref{fig-model-cycle} shows the CycleGAN
\cite{zhu2017unpaired}, which uses a generator $G$ for the mapping $X\rightarrow Y$ and another generator $F$ for $Y\rightarrow X$. Two associated adversarial
discriminators, $D_X$ and $D_Y$, are used to measure the quality of generated samples in the corresponding domains. Figure~\ref{fig:forward}
contains the forward cycle-consistency path: $x\rightarrow G(x) \rightarrow F(G(x)) \approx x$, and Figure~\ref{fig:back} is the
backward cycle-consistency path: $y\rightarrow F(y)\rightarrow G(F(y))\approx y$. The cycle consistency loss captures the intuition that if
we translate from one domain to the other and back again, we should be able to reconstruct the original input. 
However, the generated image by CycleGAN
is
deterministic. As will be shown in Section \ref{sec-exp}, these models cannot model additional variations even by adding random noise to the
inputs.

Another recent model is the UNIT \cite{liu2017unsupervised}, which performs image translation by using a shared latent space to encode
the two domains. Although they can generate multiple images via the use of a stochastic variable, the variations generated are still limited.



\section{The Proposed Model}
\label{sec-model}
Let $\mathcal A$ and $\mathcal B$ be two image domains. In supervised image-to-image translation, a sample pair $(A, B)$ is drawn from the joint distribution $P_{\mathcal A, \mathcal B}$. In this paper, we focus on unsupervised image-to-image translation, in which samples are drawn from the marginal distributions $P_{\mathcal A}$ and $P_{\mathcal B}$.

\subsection{Generators and Cycle Consistency Loss} 

\begin{figure}[ht]
	\centering
	\subfigure[X path. \label{fig:x}]
	{\includegraphics[width=0.4\textwidth]{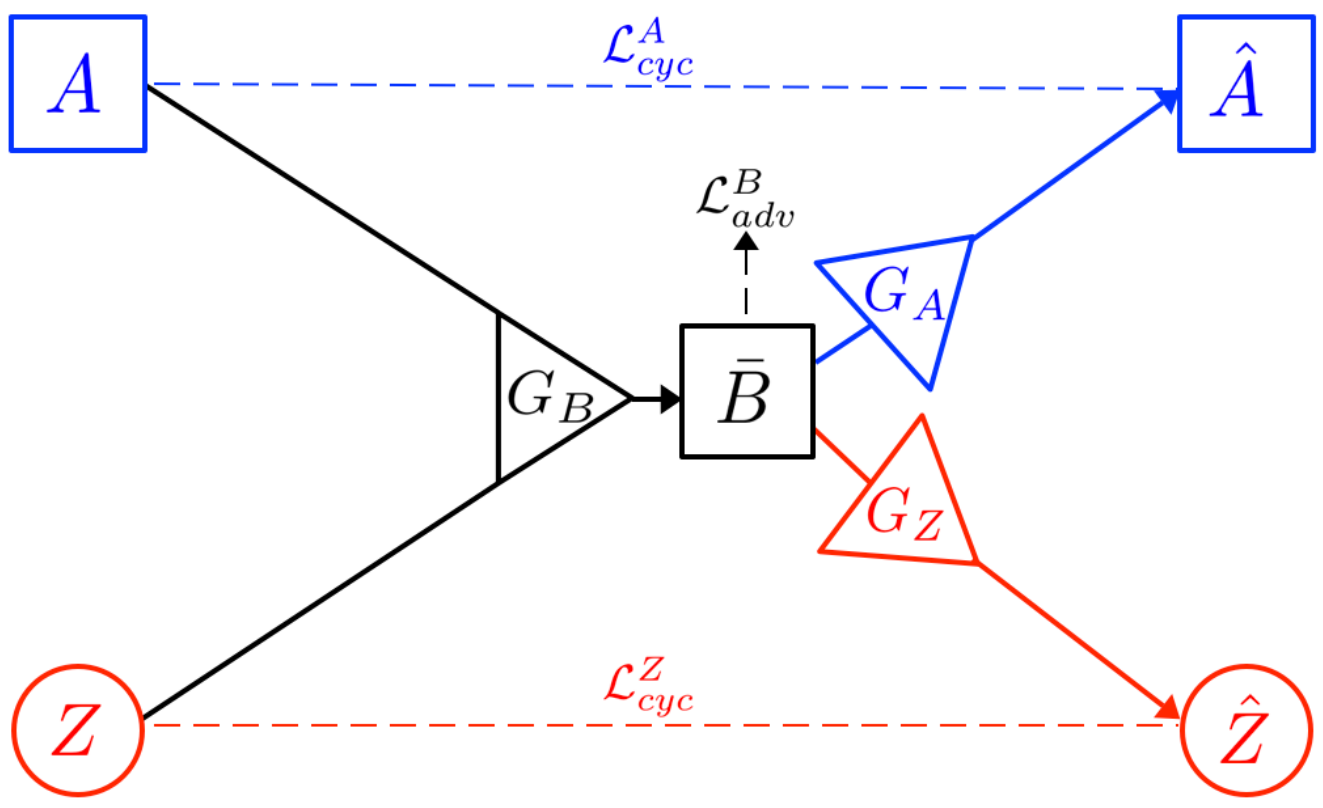}}

	\subfigure[O path. \label{fig:o}]
	{\includegraphics[width=0.4\textwidth]{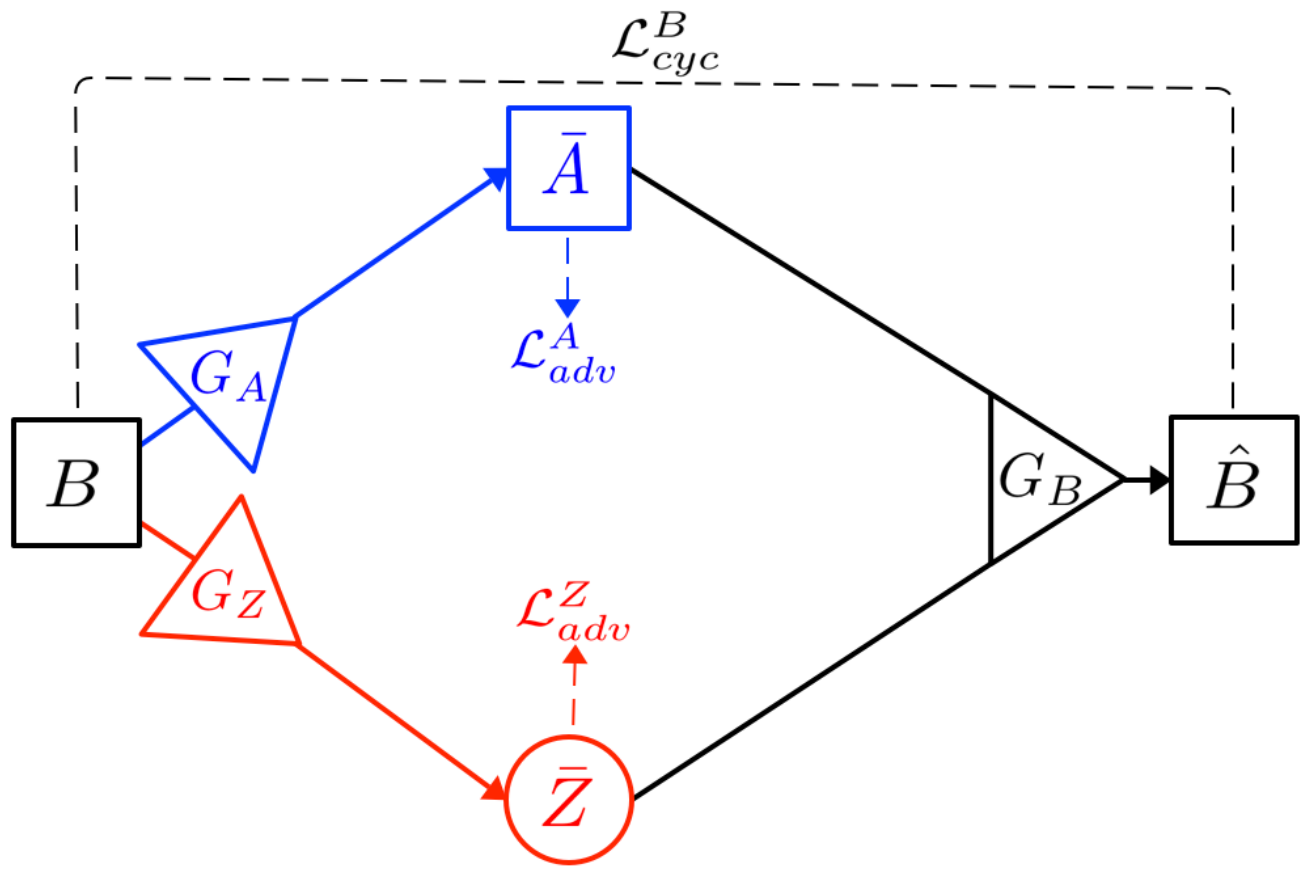}}
	\caption{The three generators in the proposed XOGAN model. The path in blue is for domain $\mathcal A$, green is for $\mathcal B$, and yellow is for $Z$.}
	\label{fig-model-gen}
\end{figure}

The proposed XOGAN model contains three generators $G_A, G_B$ and $G_Z$ (with parameters $\theta_{G_A}$, $\theta_{G_B}$ and $\theta_{G_Z}$, respectively). In this paper, we propose to use an additional variable $Z$ to model the variation when translating from domain $\mathcal A$ to domain $\mathcal B$. Given a sample $A$ drawn from $P_{\mathcal A}$ and $Z$ from the prior distribution $P_{Z}$, a fake sample $\bar{B}$ in domain $\mathcal B$ is generated by $G_B$ as
\[ \bar{B} = G_B(A, Z).\]
Given $\bar{B}$, generator $G_A$ generates a reconstruction $\hat{A}$ of $A$ in domain $\mathcal A$, and generator $G_Z$ encodes a reconstruction $\hat{Z}$ of $Z$:
\[\hat{A} = G_A(\bar{B}), \;\; \hat{Z}=G_Z(\bar B). \]
Together, this forms the X-path in Figure~\ref{fig:x}. To ensure cycle consistency, the generated sample $\bar{B}$ should contain sufficient information to reconstruct $A$ (for the path $A\rightarrow\bar{B}\rightarrow\hat{A}$), and similarly $\hat Z$ should be similar to $Z$ (for the path $Z\rightarrow\bar{B}\rightarrow \hat{Z}$).

On the other hand, given a sample $B$ in domain $\mathcal B$, generator $G_A$ can use it to generate a fake sample $\bar{A} = G_A(B)$ in domain $\mathcal A$; and generator $G_Z$ can use it to encode a fake $\bar{Z}=G_Z(B)$. Using both $\bar{A}$ and $\bar{Z}$, generator $G_B$ can recover a reconstruction of $B$ as $\hat{B}=G_B(\bar{A}, \bar{Z})$. This forms the O-path in Figure \ref{fig:o}. Again, for cycle consistency, $\hat B$ should be close to $B$.

Combining the above, the cycle consistency loss can thus be written as:
\begin{eqnarray}
\lefteqn{\mathcal L_{cyc}(\theta_{G_A}, \theta_{G_B}, \theta_{D_Z})} \nonumber \\
&=&||\hat{A} - A||_1 + ||\hat{B} - B||_1 + ||\hat{Z} - Z||_1. 
\label{eq-loss-recon}
\end{eqnarray}
Here, we use the $\ell_1$ norm, though other norms can also be used.

\begin{figure*}[ht]
	\centering
	\includegraphics[width=0.75\textwidth]{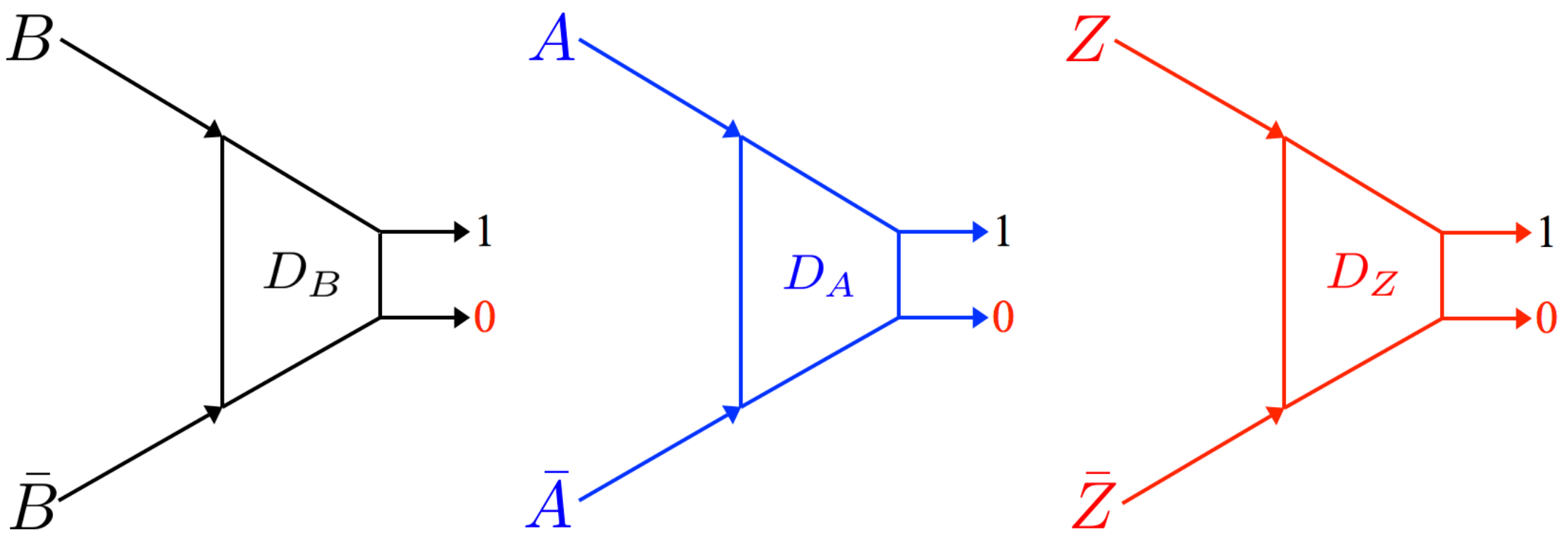}
\caption{The XOGAN discriminator. Label ``1" denotes true samples of $A, B, Z$, while label ``0" denotess the generated samples $\bar{A}, \bar{B}, \bar{Z}$.}
	\label{fig-model-dis}
\end{figure*}




\subsection{Domain Adversarial Loss} 

Minimizing (\ref{eq-loss-recon}) alone cannot guarantee that the generated fake samples $\bar{A}, \bar{B}$ and encoded variable $\bar{Z}$ follow the distributions of $P_{\mathcal A}, P_{\mathcal B}$ and $P_{Z}$. The GAN \cite{goodfellow2014generative}, which is known to be able to learn good generative models, can also be regarded as performing distribution matching. In the following, with the use of the adversarial loss, we will try to match the generated distributions $P_{G_A}$, $P_{G_B}$ and $P_{G_Z}$ with the corresponding $P_{\mathcal A}, P_{\mathcal B}$ and $P_{Z}$. 

Figure \ref{fig-model-dis} shows the three discriminators $D_A$, $D_B$ and $D_Z$ (with parameters $\theta_{D_A}, \theta_{D_B}$ and $\theta_{D_Z}$, respectively) that are used to discriminate the generated $\bar{A}, \bar{B}, \bar{Z}$ from the true $A, B, Z$. As in DiscoGAN \cite{kim2017learning} and CycleGAN \cite{zhu2017unpaired}, the discriminators are binary classifiers, and the discriminator losses are:
\begin{align}
\mathcal L_{dis}(\theta_{D_A}) =& -\mathbb E_{A\in P_{\mathcal A}} \left[\log D_A( A)\right]  \nonumber \\
&- \mathbb E_{\bar{A}\in P_{G_A}}\left[\log \left(1-D_A(\bar{ A})\right)\right]  \label{loss-dis-a}, \\
\mathcal L_{dis}(\theta_{D_B}) =& -\mathbb E_{B\in P_{\mathcal B}} \left[\log D_B( B)\right]  \nonumber \\
&- \mathbb E_{\bar{B}\in P_{G_B}}\left[\log \left(1-D_B(\bar{ B})\right)\right]  \label{loss-dis-b}, \\
\mathcal L_{dis}(\theta_{D_Z}) =& -\mathbb E_{Z\in P_{Z}} \left[\log D_Z( Z)\right]  \nonumber \\
&- \mathbb E_{\bar{Z}\in P_{G_Z}}\left[\log \left(1-D_Z(\bar{ Z})\right)\right]  \label{loss-dis-z}.
\end{align}

In GAN, the generators, besides trying to minimize the cycle consistency loss, also need to confuse their corresponding discriminators. The adversarial losses for the generators are 
\begin{eqnarray*}
\mathcal L_{adv}(\theta_{G_{A}}) &=& -\mathbb E_{A\in P_{\mathcal A}}\left[\log D_A(\bar{ A})\right],\\
\mathcal L_{adv}(\theta_{G_{B}}) &=& -\mathbb E_{B\in P_{\mathcal B}}\left[\log D_B(\bar{ B})\right],\\
\mathcal L_{adv}(\theta_{G_{Z}}) &=& -\mathbb E_{Z\in P_{Z}}\left[\log D_Z(\bar{ Z})\right].\\
\end{eqnarray*}


To ensure both cycle consistency and distribution matching, the total loss for the generators is a combination of the cycle consistency loss in (\ref{eq-loss-recon}) and the adversarial losses:
\begin{align}
\mathcal L_{gen}(\theta_{G_A}, \theta_{G_B}, \theta_{D_Z}) =& \mathcal L_{adv}(\theta_{G_{A}}) + \mathcal L_{adv}(\theta_{G_{B}}) + \mathcal L_{adv}(\theta_{G_{Z}}) \nonumber \\
+&\lambda \mathcal L_{cyc}(\theta_{G_A}, \theta_{G_B}, \theta_{G_Z}),
\label{eq-loss-total}
\end{align}
where $\lambda$ controls the balance between the two types of losses. In the experiment, $\lambda$ is set to 10.

\subsection{Training Procedure}

In each iteration, we sample a mini-batch of images $A$'s and $B$'s from $P_{\mathcal A}$, $P_{\mathcal B}$, and variable $Z$ from prior distribution $P_{Z}$ (which is the standard normal distribution $\mathcal N(0, I)$). They are fed through the X- and O-paths in Figure \ref{fig-model-gen} to obtain the generated samples $\bar{A}$, $\bar{B}$, $\bar{Z}$, and the reconstructed samples $\hat{A}$, $\hat{B}$, $\hat{Z}$. The real and generated samples are then input to the three discriminators.

As in GAN, we minimize the generators' objective (\ref{eq-loss-total}) for $k$ steps and the discriminators' objectives ((\ref{loss-dis-a}),
(\ref{loss-dis-b}) and (\ref{loss-dis-z})) for one step. We use the ADAM optimizer \cite{kingma2014adam} with learning rate 0.0002.

\section{Experiments}
\label{sec-exp}
In this section, experiments are performed on a number of commonly used data sets for image translation.

\begin{enumerate}
\item 
{\sf edges2shoes} and {\sf edges2handbags}:\footnote{\url{https://people.eecs.berkeley.edu/~tinghuiz/projects/pix2pix/data sets/}}
These two data sets  
have been used in \cite{isola2016image}.
The {\sf edges2shoes} data set contains about 50k paired images, and the {\sf edges2handbags} data set contains about 140k paired images. In both data sets, domain $\mathcal A$ contains edge images and domain $\mathcal B$ contains real objects (shoes and bags). Note that one real object can be mapped to only one edge map, but an edge map can correspond to multiple objects. 
	
Although the two data sets contain paired images, we separate the paired sets by sampling domain $\mathcal A$ images in the first half pairs, and domain $\mathcal B$ images in the other half pairs. Hence, there is no paired data in the training set, and the task is unsupervised image-to-image translation.
	
\item {\sf CelebA}:\footnote{\url{http://mmlab.ie.cuhk.edu.hk/projects/CelebA.html}} 
This is a large-scale face attributes data set with more than 200K celebrity images, each with 40 attribute annotations
\cite{liu2015deep}.
In the experiment, we use the hair color attribute. Domain $\mathcal A$ contains faces with black hair, and domain $\mathcal B$ contains faces
with other hair colors. As the hair in many male {\sf CelebA} faces is not apparent, we only use the female faces.	
\end{enumerate}
All the input images are rescaled to $128\times 128$.

In the proposed XOGAN, we use the U-Net \cite{ronneberger2015u}, which adds skip connections between mirrored layers of the 7 downsampling layers
and the 7 upsampling layers, for the image generators $G_A$ and $G_B$. For $Z$, the generator (or encoder) $G_Z$ is a 7-layer strided
convolutional network with residual blocks \cite{he2016deep}. We set its dimensionality to 8 for the {\sf edges2shoes} and {\sf edges2handbags} data sets,
respectively, and 4 for the {\sf CelebA} data set. For the image discriminators $D_A$ and $D_B$, we use the patch-discriminator in \cite{li2016precomputed}, which only classifies input images at the scale of patches, with $70\times70$ overlapping patches. The discriminator $D_Z$ is a simple two-layer multi-layer-perception. As in \cite{zhu2017unpaired}, the hyperparameter $\lambda$ in (\ref{eq-loss-total}) is set to 10. In each training iteration, we update the generators twice and then update the discriminators once.

Existing  unsupervised image-to-image translation models such as DiscoGAN \cite{kim2017learning} and CycleGAN \cite{zhu2017unpaired} can only generate one target image. Instead, we use the following setup as the baseline model:
\begin{enumerate}
	\item Noisy DiscoGAN: This is a variant of DiscoGAN. It uses the same generators as $G_A, G_B$ in XOGAN, but the generator from domain $\mathcal A$ to domain $\mathcal B$ is augmented with random Gaussian noise. This allows the generation of different images in domain $\mathcal B$ given the same image from domain $\mathcal A$. We do not compare with CycleGAN \cite{zhu2017unpaired} and DualGAN \cite{yi2017dualgan}, as they are very similar to DiscoGAN.
	\item UNIT \cite{liu2017unsupervised}: The UNIT model uses two variational autoencoders \cite{kingma2013auto} with shared latent space as
	cross-domain image translators.  It also uses cycle consistency for unpaired image-to-image translation. For each input image, we sample multiple latent codes $z$'s, and use them to generate different outputs.
\end{enumerate}

\subsection{Translating $\mathcal A$ to $\mathcal B$ with Random $Z$}
\label{sec-AB}

\begin{figure}[ht]
	\centering
	\subfigure[Edges2Shoes.]
	{\includegraphics[width=0.45\textwidth]{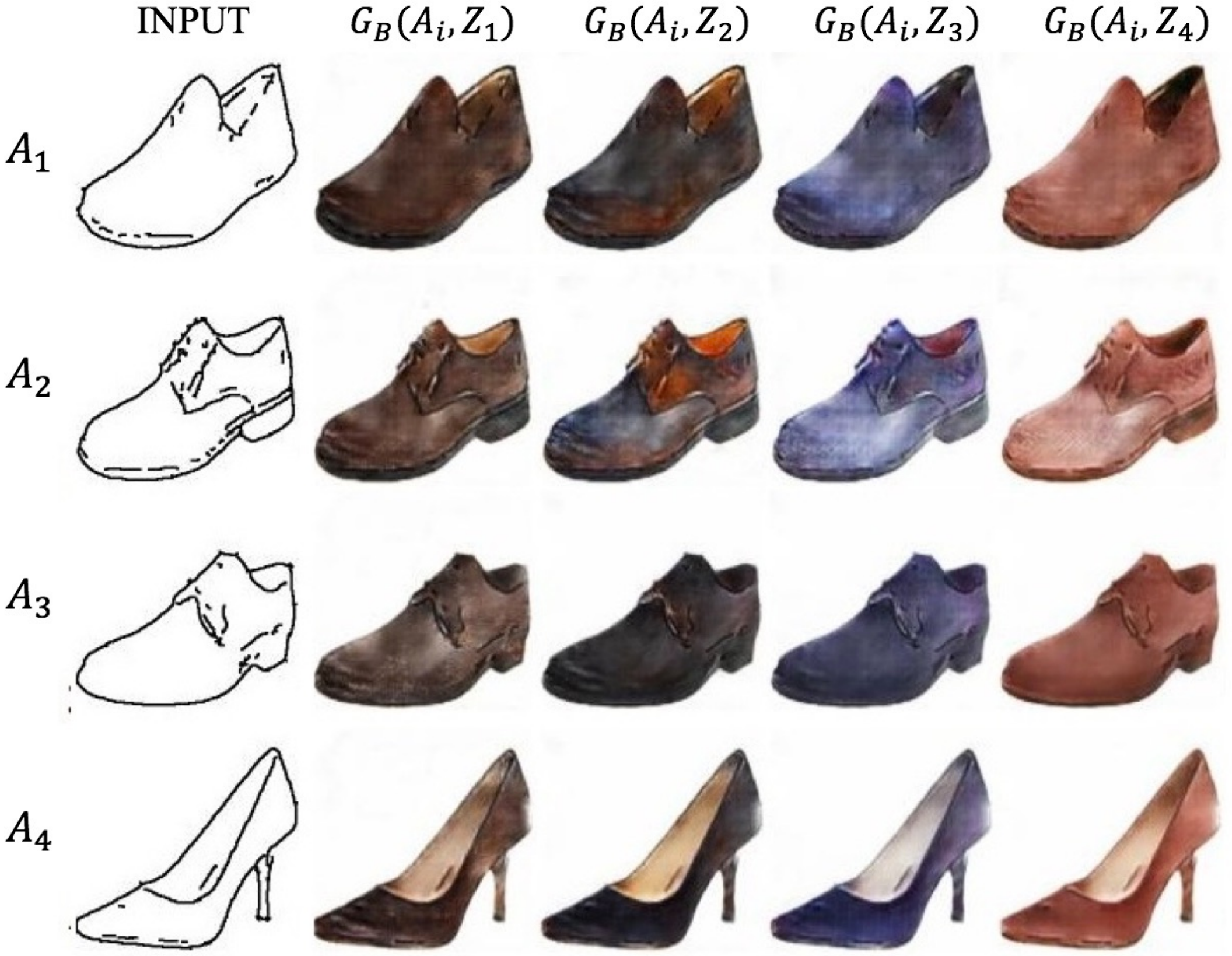}}

	\subfigure[Edges2Handbags.]
	{\includegraphics[width=0.45\textwidth]{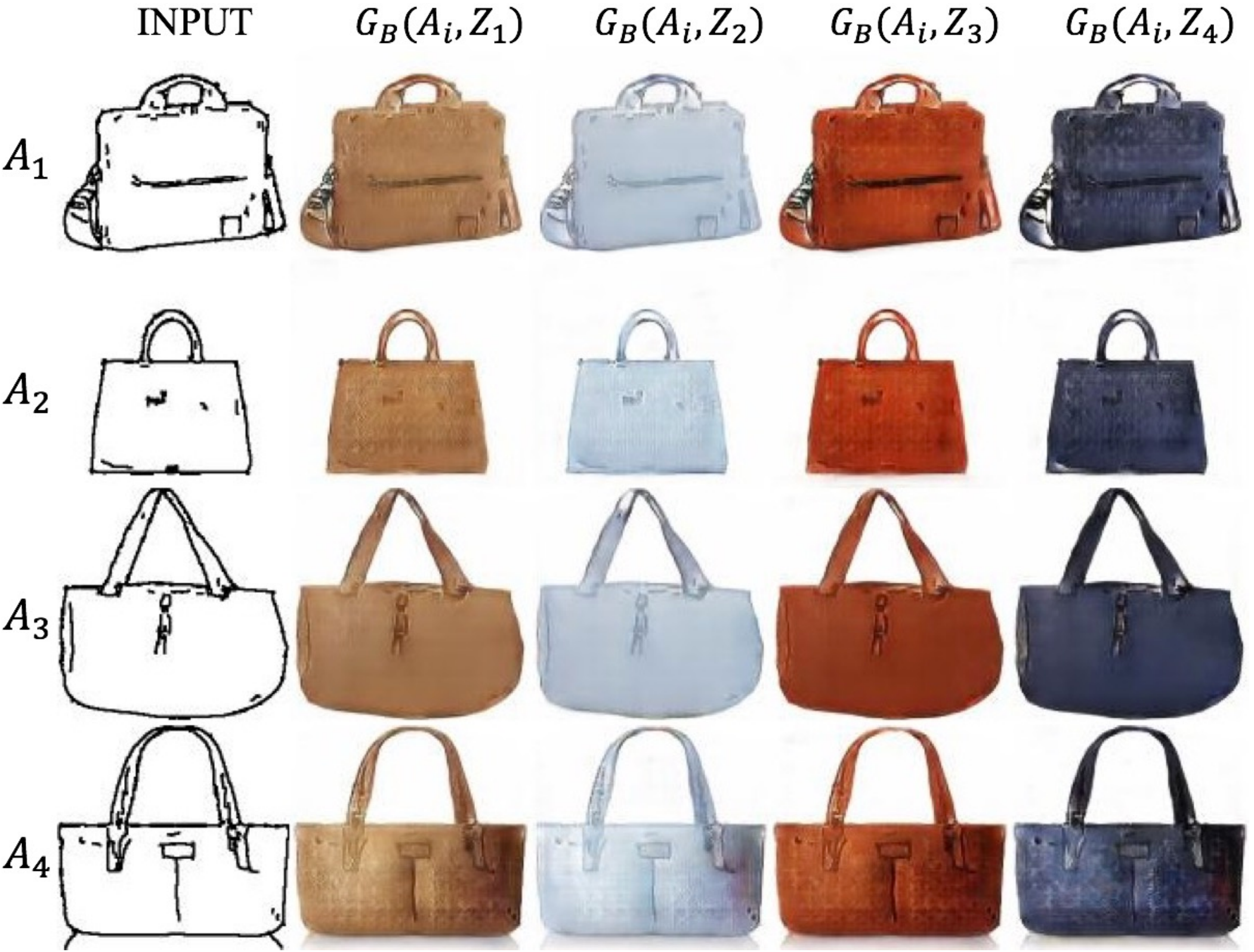}}
	\caption{Edges to shoes and handbags experiment. The right 4 images in each row are translated from input image $A_i$ to $\bar{B}_i$ with
	random variable $Z_j, j=1,\dots,4$. As can be seen, the colors are consistent in the right 4 columns.}
	\label{fig-edges2objects}
\end{figure}

To show the consistency of the learned additional variables, we sample different random variables $Z_j$'s to generate different fake images
$\bar{B}_{ij}=G_B(A_i, Z_j)$ for each input image $A_i$. We keep the random variable set $\{Z_j\}$ the same for different $A_i$'s, so that when
$Z_j$ or $j$ is fixed, the generated samples $\bar{B}_{ij}$ should have some fixed attributes when $i$ is changing, such as the color as shown in
Figures~\ref{fig-edges2objects} and \ref{fig-haircolor}. 

In this experiment, we randomly sample 4 input images $\{A_i, i=1,\dots,4\}$  in domain $\mathcal A$ from the test set. For each $A_i$, we generate 4 images $\{\bar{B}_{ij} = G_B(A_i, Z_j), j=1,\dots,4\}$ using different $Z_j$'s sampled from the standard normal distribution. 

\subsubsection{{\sf edges2shoes} and {\sf edges2handbags} Data Sets}



Figure \ref{fig-edges2objects} shows that the proposed XOGAN can generate realistic photos of shoes and handbags. 
With the help of the cycle consistency loss on variable $Z$, the generated objects have the same color for the same $Z_j$. This makes it possible
to generate plausible and colorful objects by just drawing edges like those in domain $\mathcal A$. Besides, we can also control the color of
generated images by $Z$. Note that our task is different from the drawing softwares in two ways. (i) We show that the end-to-end deep neural network is able to generate plausible and diverse objects given edges; (ii) The generated objects do not only fill in the colors, but also have smooth textures that make them look real. 

\begin{figure}[ht]
	\centering
	\subfigure[Edges2Shoes.]
	{\includegraphics[width=0.45\textwidth]{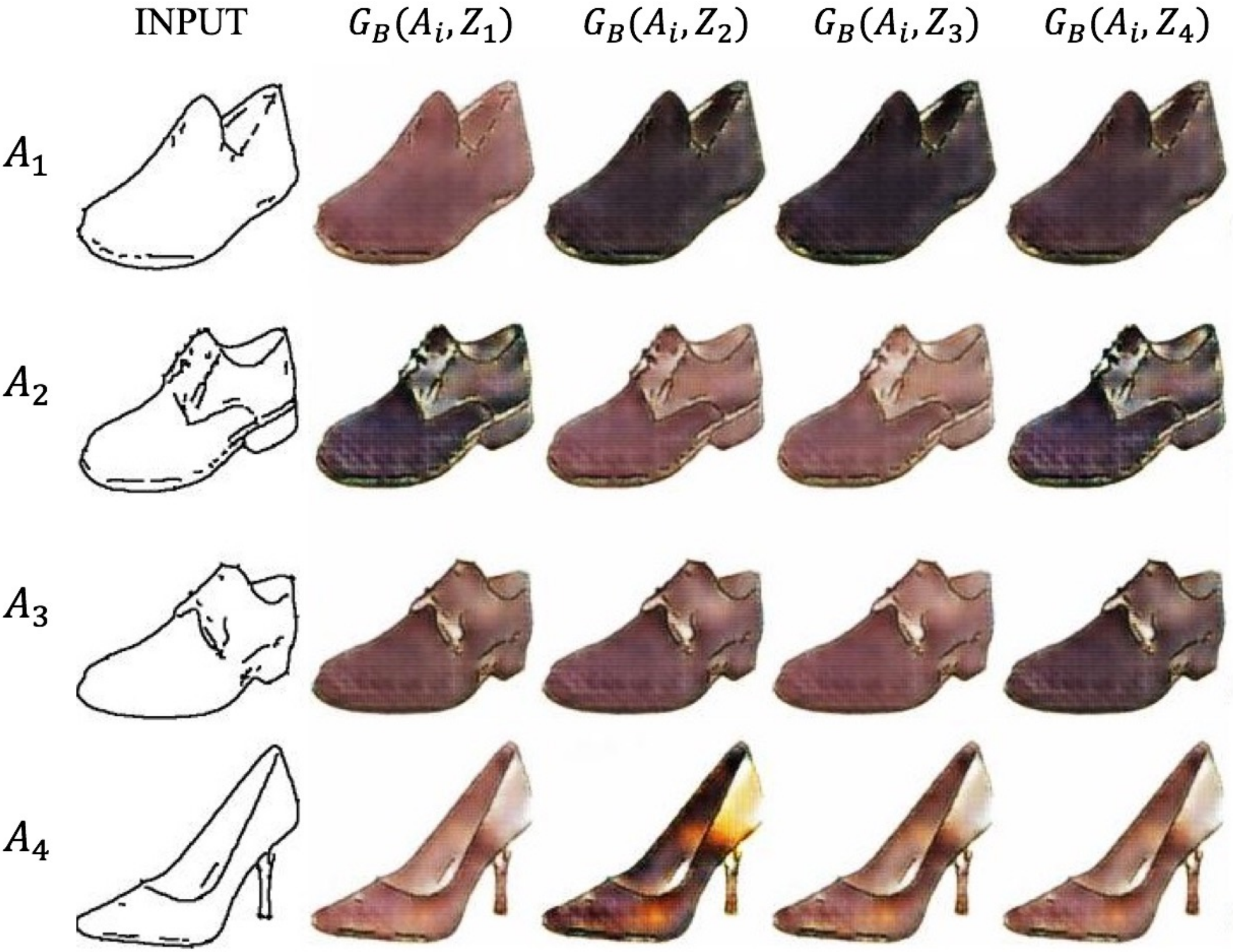}}

	\subfigure[Edges2Handbags.]
	{\includegraphics[width=0.45\textwidth]{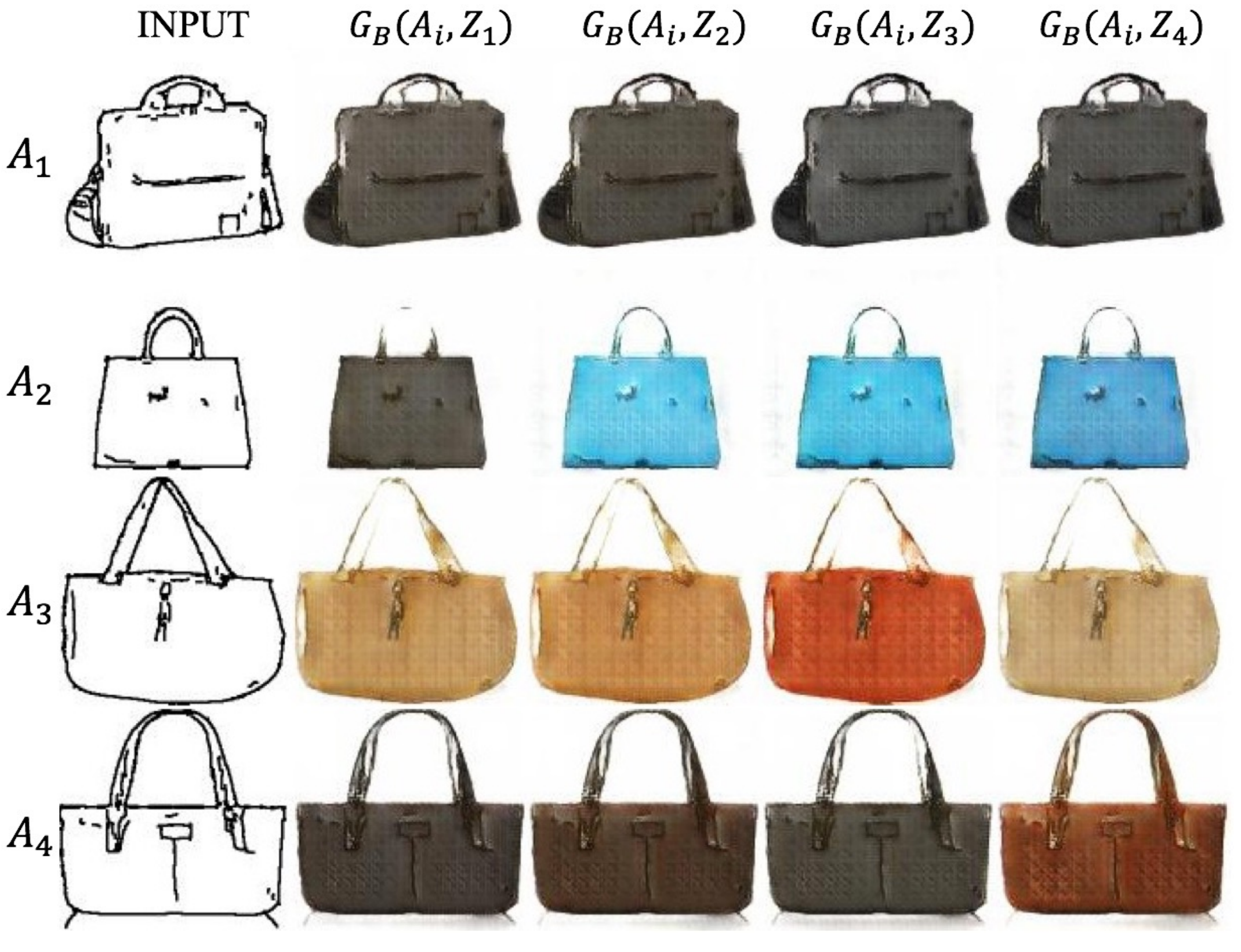}}
	\caption{Edges to shoes and handbags experiment of noisy DiscoGAN. The right 4 images in each row are translated from input image $A_i$ to $\bar{B}_i$ with random variable $Z_j, j=1..4$.}
	\label{fig-edges2objects-base}
\end{figure}

Figure \ref{fig-edges2objects-base} shows the results obtained by the noisy DiscoGAN. Although multiple images can be generated, they are
inferior as compared to those obtained with the proposed model in the following ways. (i) The images $\{G_B(A_i, Z_j)\}$ generated by the noisy
DiscoGAN are not as good as those of  XOGAN; (ii) The images generated by the noisy DiscoGAN are not as diverse as the proposed XOGAN, i.e.,
$G_B(A_i, Z_j)$ has similar color for different $j$'s in each row. This may be due to the  mode collapse problem of GAN \cite{goodfellow2014generative}, which means that it tends to generate images from a single mode (in this case, color).

\begin{figure}[ht]
	\centering
	\subfigure[Edges2Shoes.]
	{\includegraphics[width=0.45\textwidth]{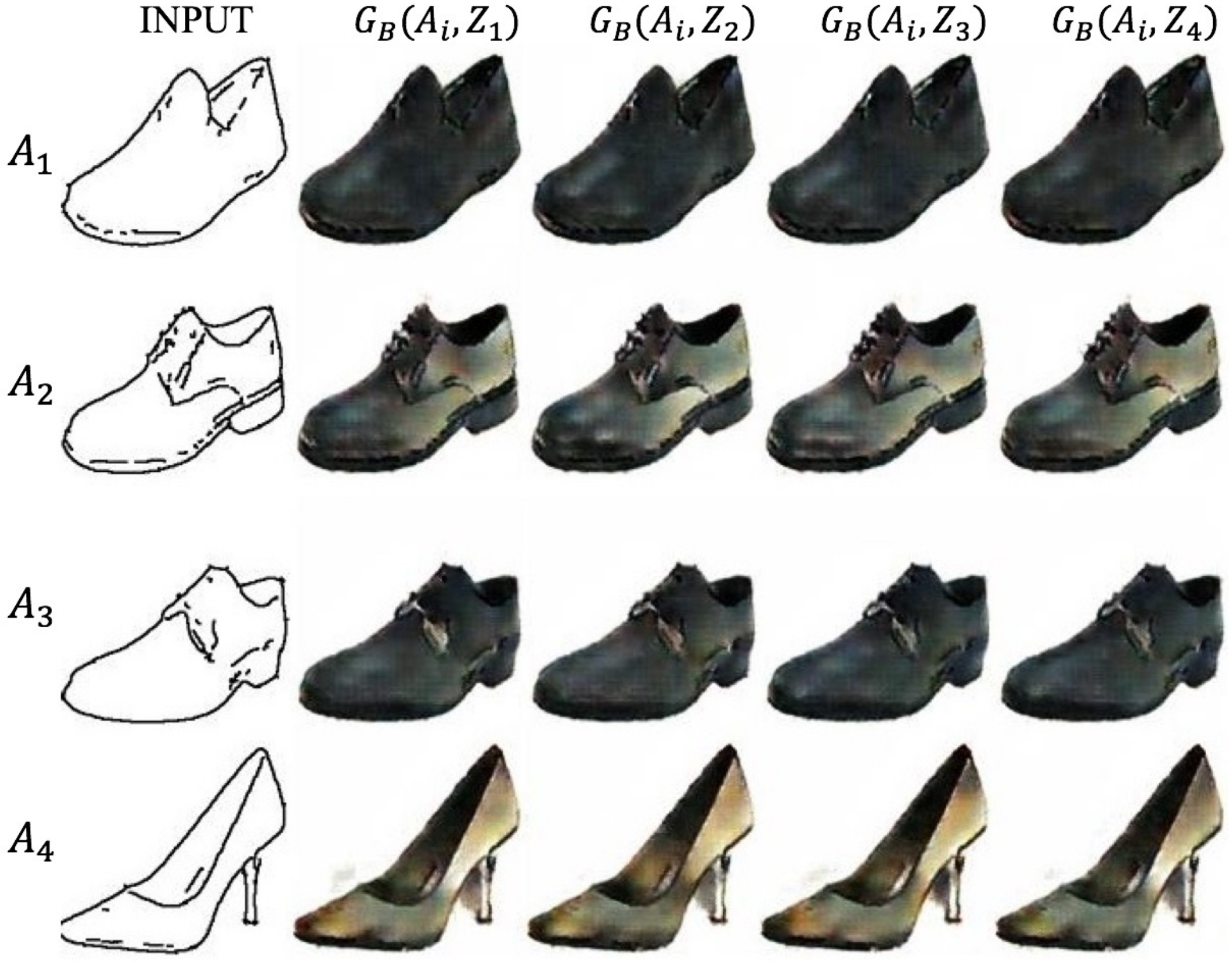}}

	\subfigure[Edges2Handbags.]
	{\includegraphics[width=0.45\textwidth]{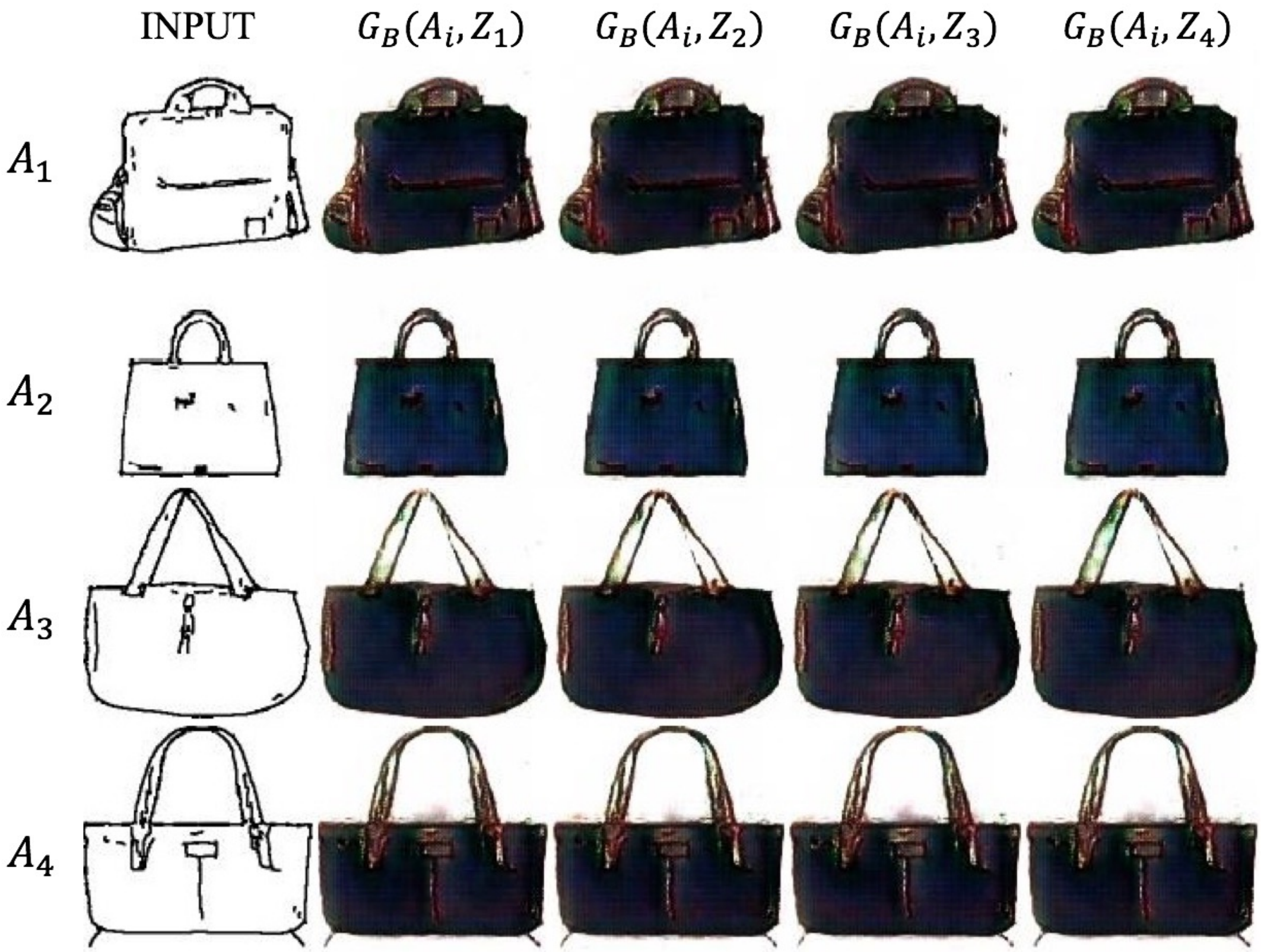}}
	\caption{Edges to shoes and handbags experiment of \textit{UNIT} model. The right 4 images in each row are translated from input image $A_i$ to $\bar{B}_i$ with random variable $Z_j,  j=1,\dots,4$.}
	\label{fig-edges2objects-unit}
\end{figure}

Figure \ref{fig-edges2objects-unit} shows the results obtained by UNIT. Similar to the noisy DiscoGAN, almost all the generated images tend to
have the same color. From our observations, the generated samples of UNIT also suffer from mode collapse during training. The sampled latent
variables do not produce diverse outputs in each row. 

The above results show that the proposed XO-path is essential in both generating plausible and consistency target images. From the {\sf
edges2shoes} and {\sf edges2handbags} data sets, we show that XOGAN is able to generate plausible images in the target domain $\mathcal B$.
Different from the baseline models of noisy DiscoGAN and UNIT, we can sample mulitple and consistent output images with different $Z_j$'s.
Besides, the results of XOGAN are more diverse with the help of $Z$. 

\begin{figure}[ht]
	\centering
	{\includegraphics[width=0.45\textwidth]{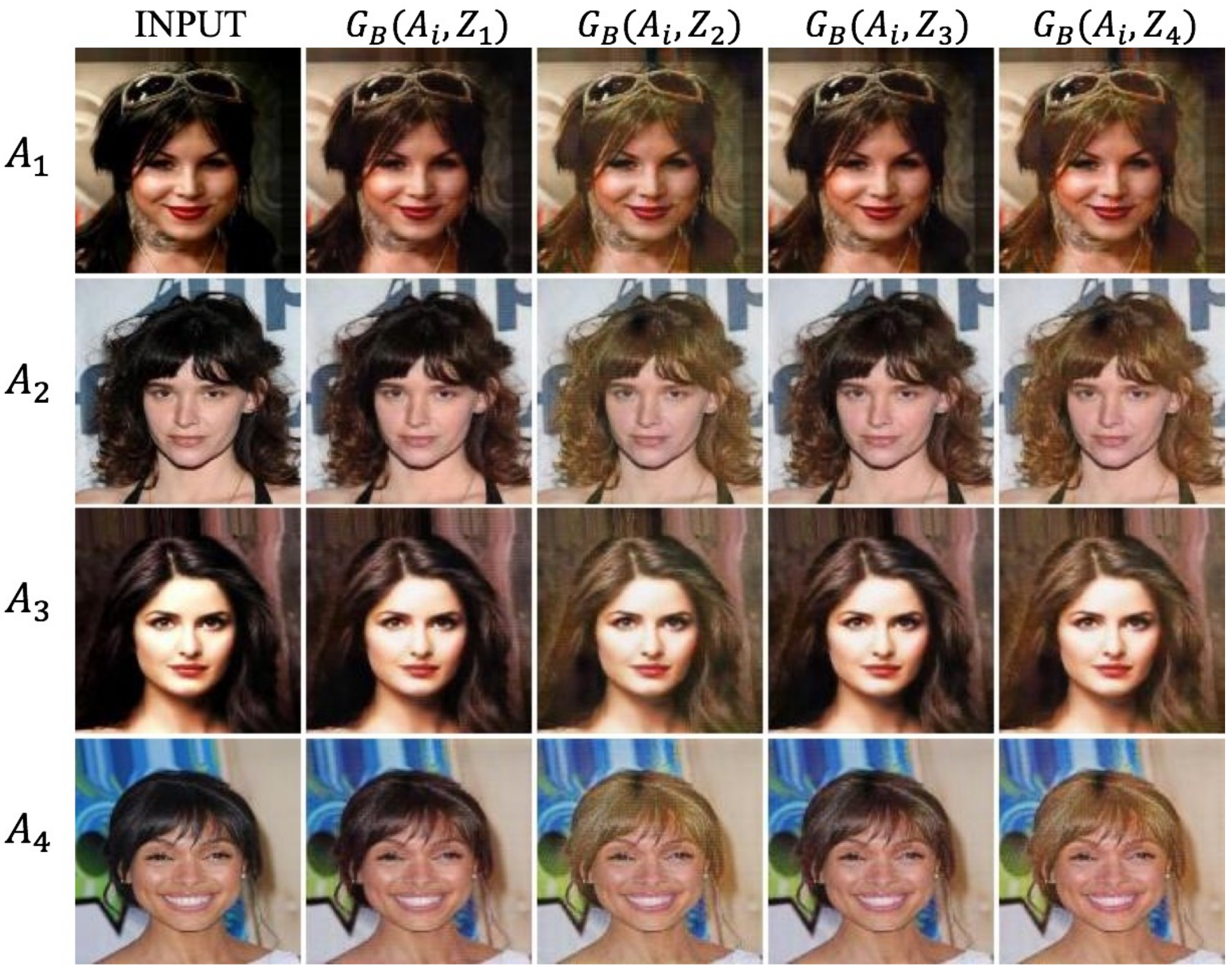}}
	\caption{CelebA hair color conversion experiment. We transfer the black hair faces to faces with other hair colors. The hair colors are
	consistent for different $G_B(A_i, Z_j)$, when $j$ is fixed.}
	\label{fig-haircolor}
\end{figure}

\subsubsection{{\sf CelebA} Data Set} 

Figure \ref{fig-haircolor} shows the generated faces with different hair colors. 
Variable $Z$ is used to model the different hair colors in domain $\mathcal B$. 
Note that the other parts in the images are almost unchanged. Hence, the color change is focused on specific parts (hairs here).
 

\subsection{Translating $\mathcal B$ to $\mathcal A$ to $\mathcal B$ with Substituted $Z$}
\label{sec-BAB}

In this section, we study whether $Z$ can encode relevant information in domain $\mathcal B$ by substituting $\bar{Z}_j$ among different images in domain $\mathcal B$.

In this experiment, we randomly sample 4 input images $\{B_i, i=1,\dots,4\}$ in domain $\mathcal B$ from the test set. For each $B_i$, we
generate 4 corresponding images $\{\bar{A}_i=G_A(B_i), i=1,\dots,4\}$ in domain $\mathcal A$ and encode its additional variation in
$\{\bar{Z}_i=G_Z(B_i)\}$. As in the previous {\sf edges2shoes} experiment, $B_i$ represents the colored shoes, $\bar{A}_i$ is its corresponding
edge image and $\bar{Z}_i$ should encode content inside the edge. We concatenate $\bar{A}_i$ with different $\bar{Z}_j$'s to generate various images $\{\hat{B}_{ij}=G_B(\bar{A}_i, \bar{Z}_j), i=1,\dots,4, j=1,\dots,4$ in domain $\mathcal B$.


\subsubsection{{\sf edges2shoes} and {\sf edges2handbags} Data Sets} 
\begin{figure}[ht]
	\centering
	\subfigure[Edges2Shoes.]
	{\includegraphics[width=0.45\textwidth]{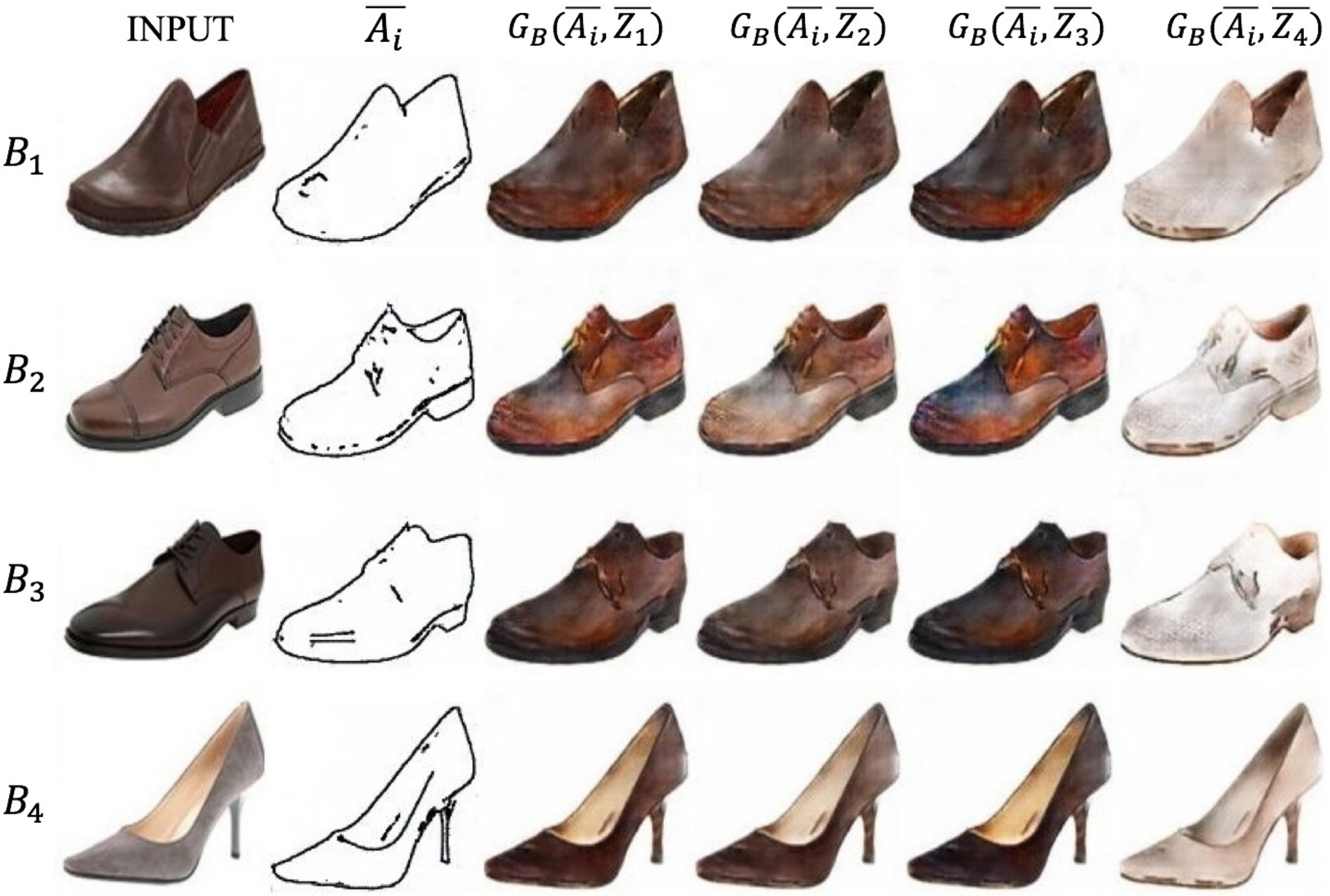}}

	\subfigure[Edges2Handbags.]
	{\includegraphics[width=0.45\textwidth]{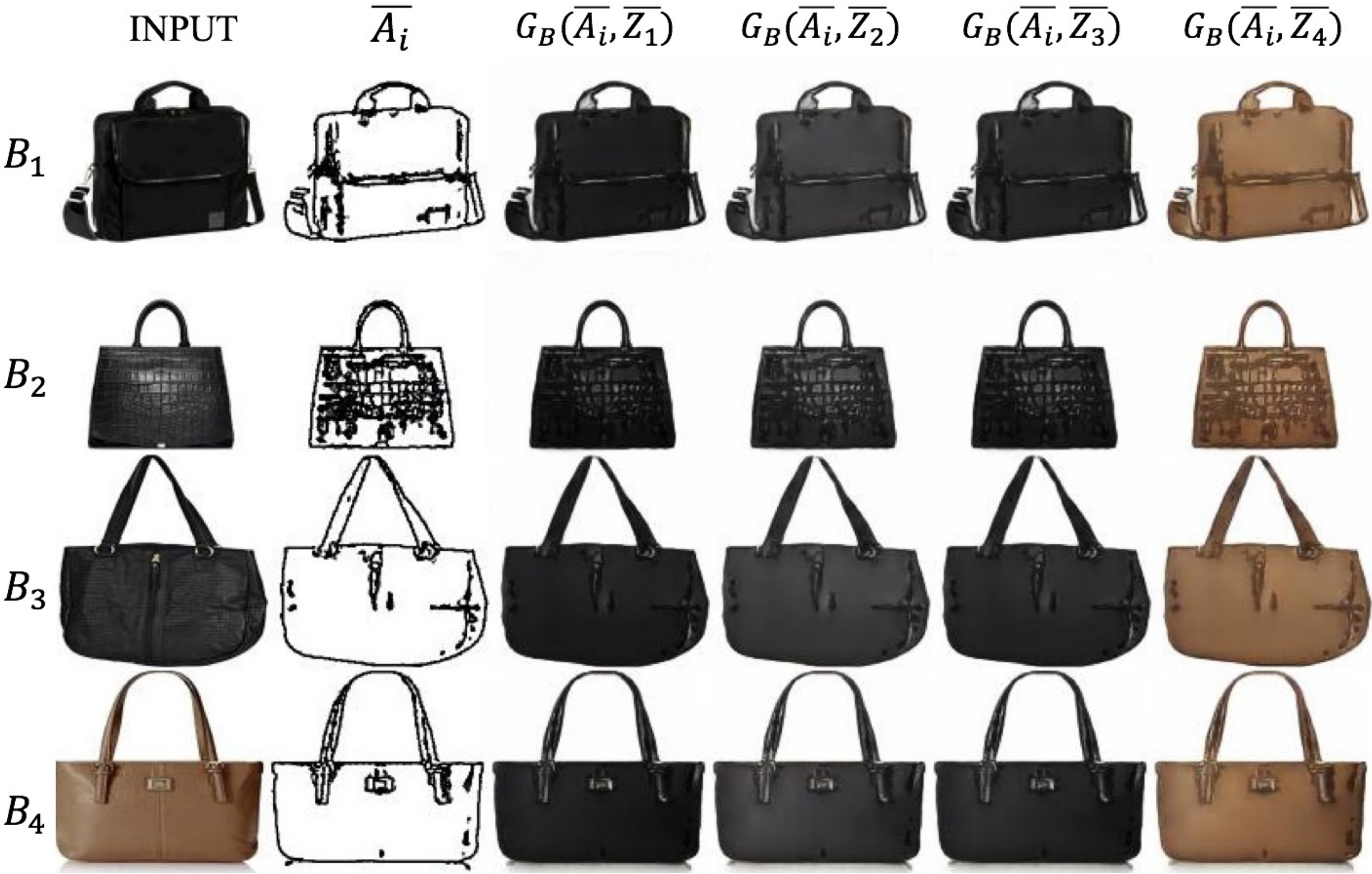}}
	\caption{Edges to shoes and handbags experiment with variable substitution. The edge image $\bar{A}_i$ is shown in the second column. The rightmost 4 images are translated with the concatenation of $\bar{A}_i$ with variation $\bar{Z}_j$ for each input image $B_j$. The color is replaced for different inputs.}
	\label{fig-edges2objects-2}
\end{figure}

To see what the additional variable $Z$ encodes, we substitute different
$Z_j$'s given different colored shoes and handbags. As shown in Figure \ref{fig-edges2objects-2}, we can modify images in domain $\mathcal B$. In
these two pictures, we show the input shoes and handbags $B_i$ in the first column. We first generate the corresponding edge image  $\bar{A}_i$,
shown in the second column of Figure \ref{fig-edges2objects-2}, of each $B_i$ using generator $G_A$. The generated edges describe the contour of
the given inputs well. Considering that edge images ignore the content of the given shoes or handbags, we encode the content or color into the
additional variable $\{\bar{Z}_i=G_Z(B_i), i=1,\dots,4\}$. As we assumed, $Z_i$ should contain the corresponding color information of input image
$B_i$. In order to show the relevance between $Z_i$ and color of $B_i$, we concatenate each edge image $\bar{A}_i$ with different $\bar{Z}_j$'s
to generate new images in domain $\mathcal B$ as $\hat{B}_{ij}=G_B(\bar{A}_i, \bar{Z}_j)$. In the first row of Figure \ref{fig-edges2objects-2},
we show $B_1$ and $\bar{A}_1$ in nthe first two columns, and the last four columns show $\hat{B}_{1j}, j=1,\dots,4$. Since $\hat{B}_{1j}$ is
generated from the concatenation of the edge image $\bar{A}_1$ and different codes $\bar{Z}_j$, $\hat{B}_{1j}$ should have the same shape as
$B_1$ and the same color as $B_j, j=1,\dots,4$. 

In the figure, we can see that the shape is consistent in each rows and colors are consistent in each of the four rightmost columns. More
importantly, when $j\in[1,\dots,4]$ is fixed, all images $\hat{B}_{ij}$'s in that column have similar colors as the input image $B_j$. As for the
{\sf edges2shoes} results in Figure~\ref{fig-edges2objects-2}, the shoes in the last column have similar colors as the last input shoe $B_4$. It
verifies our assumption that $\bar{Z}$ encodes relevant information of its corresponding inputs $B$. This is interesting that, we can see what
our object will look like by replacing its color with that of the other objects. In real-world applications like fitting in a clothes shop, the
user does not need to try on over and over again, if they want to try the same clothes with different colors.

\subsubsection{{\sf CelebA} Hair Color Conversion} 
\begin{figure}[htbp]
	\centering
	\includegraphics[width=0.45\textwidth]{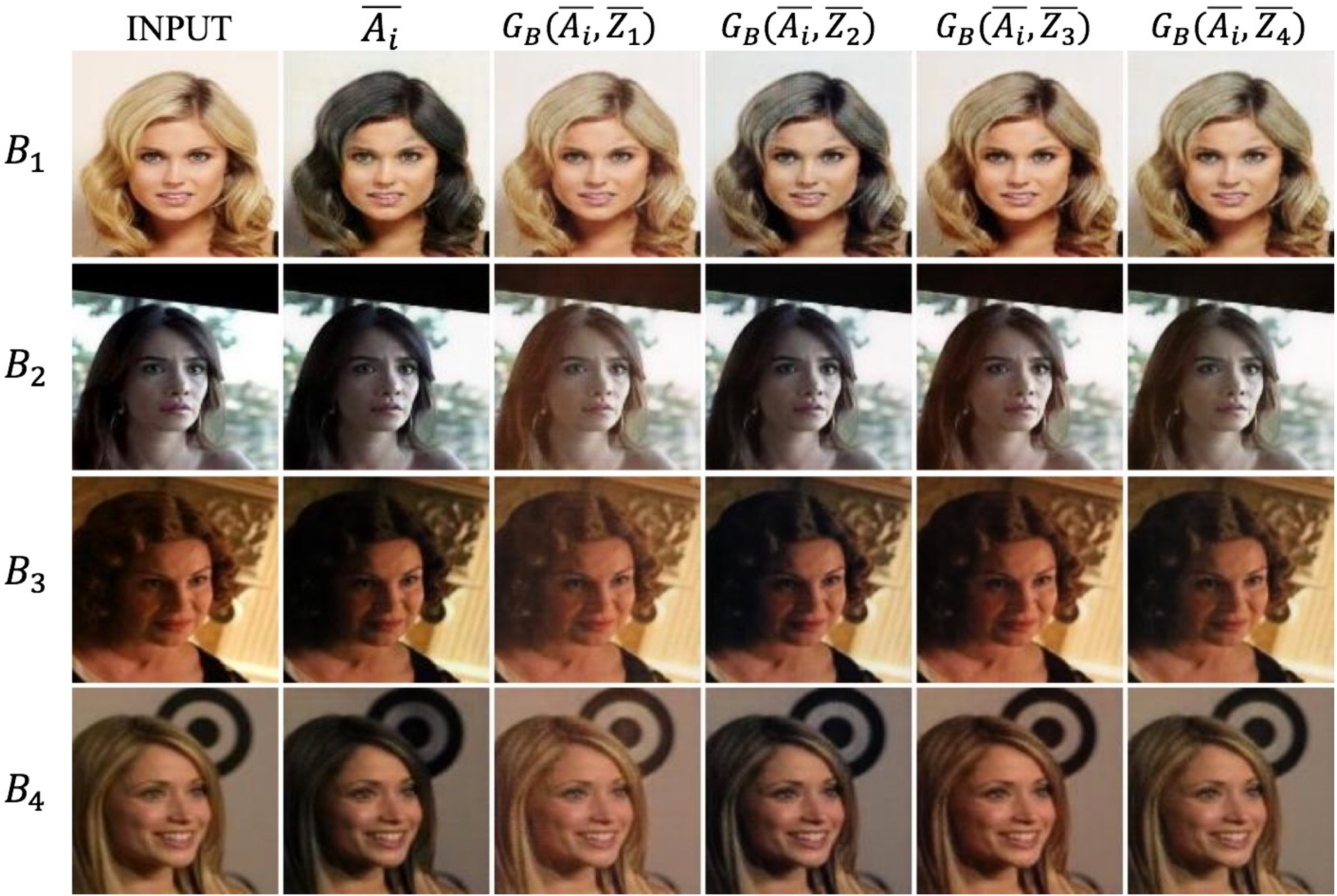}
	\caption{Hair color translation experiment. 
		The second column
		transfers non-black hair to black.
		The hair color of each person $B_i$ is encoded in $\bar{Z}_i$. By concatenating $\bar{A}_i$ with different $\bar{Z}_j, j=1..4$, we can modify the hair color.}
	\label{fig-haircolor-2}
\end{figure}

We perform the $\mathcal B$-to-$\mathcal A$-to-$\mathcal B$ path on the {\sf CelebA} data set again. Input faces are sampled from domain
$\mathcal B$ where the hair colors are not black. If the user wants to change the hair color to black, we can easily transfer the photo from
$B_i$ to $\bar{A}_i$. An interesting result is to replace the hair color with that of another person. Figure \ref{fig-haircolor-2} shows that we
can replace the blond hair to gray or brown by substituting the code $\bar{Z}_i$ with that from a gray or brown person.

In this section, we show that not only the proposed model can generate plausible and diverse images from domain $\mathcal A$ to domain $\mathcal
B$ and vice versa, but also can modify the color of specific features in domain $\mathcal B$ by substituting the additional variable $Z$. The
proposed cycle consistency constraints guarantee a good joint relationship between $\mathcal A$ and $\mathcal B$. They also encode consistent features in variable $Z$ for both random generation in Section~\ref{sec-AB} and color substitution in Section~\ref{sec-BAB}.

\section{Conclusion}
\label{sec-conclude}
In this paper, we presented a generative model called XOGAN for unsupervised image-to-image translation with additional variations in the
one-to-many translation setting. We showed that we can generate plausible images in both domains, and the generated samples are more diverse than
the baseline models. Not only does the additional variable $Z$ learned lead to more diverse results, it also controls the colors in certain parts
of the generated images. Experiments on the {\sf edges2shoes}, {\sf edges2handbags} and {\sf CelebA} data sets showed that the learned variable $Z$ is meaningful when generating images in domain $\mathcal B$.

The proposed method can be extended in several ways. First, the prior distribution of $Z$ is a standard normal distribution, which may be too
simple to model more complex variations. This can be improved with more complicated prior distributions introduced in VAE
\cite{kingma2016improved, maaloe2016auxiliary}. Second, the variations in our model are mostly related to color. We hope that our model can be
improved to change other attributes such as hair styles and ornaments.
Besides, we can also consider a many-to-many mapping based on the one-to-many framework. Similar to the summer-to-winter task in
\cite{zhu2017unpaired}, there exist many winter images corresponding to a single summer image and vice versa. Further, we can extend the proposed
model to other domains such as text or speech.  The additional variation can be different voices when translating text to speech. 


\bibliographystyle{IEEEtran}
\bibliography{ref}

\begin{thebibliography}{10}
\providecommand{\url}[1]{#1}
\csname url@samestyle\endcsname
\providecommand{\newblock}{\relax}
\providecommand{\bibinfo}[2]{#2}
\providecommand{\BIBentrySTDinterwordspacing}{\spaceskip=0pt\relax}
\providecommand{\BIBentryALTinterwordstretchfactor}{4}
\providecommand{\BIBentryALTinterwordspacing}{\spaceskip=\fontdimen2\font plus
\BIBentryALTinterwordstretchfactor\fontdimen3\font minus
  \fontdimen4\font\relax}
\providecommand{\BIBforeignlanguage}[2]{{%
\expandafter\ifx\csname l@#1\endcsname\relax
\typeout{** WARNING: IEEEtran.bst: No hyphenation pattern has been}%
\typeout{** loaded for the language `#1'. Using the pattern for}%
\typeout{** the default language instead.}%
\else
\language=\csname l@#1\endcsname
\fi
#2}}
\providecommand{\BIBdecl}{\relax}
\BIBdecl

\bibitem{isola2016image}
P.~Isola, J.-Y. Zhu, T.~Zhou, and A.~A. Efros, ``Image-to-image translation
  with conditional adversarial networks,'' Preprint arXiv: 1611.07004, 2016.

\bibitem{kim2017learning}
T.~Kim, M.~Cha, H.~Kim, J.~K. Lee, and J.~Kim, ``Learning to discover
  cross-domain relations with generative adversarial networks,'' in
  \emph{Proceedings of the 34th International Conference on Machine Learning},
  2017, pp. 1857--1865.

\bibitem{taigman2016unsupervised}
Y.~Taigman, A.~Polyak, and L.~Wolf, ``Unsupervised cross-domain image
  generation,'' Preprint arXiv:1611.02200, 2016.

\bibitem{zhu2017unpaired}
J.-Y. Zhu, T.~Park, P.~Isola, and A.~A. Efros, ``Unpaired image-to-image
  translation using cycle-consistent adversarial networks,'' in \emph{IEEE
  International Conference on Computer Vision}, 2017, pp. 2223--2232.

\bibitem{yi2017dualgan}
Z.~Yi, H.~Zhang, P.~Tan, and M.~Gong, ``Dualgan: unsupervised dual learning for
  image-to-image translation,'' in \emph{IEEE International Conference on
  Computer Vision}, 2017, pp. 2849--2857.

\bibitem{ledig2017photo}
C.~Ledig, L.~Theis, F.~Huszar, J.~Caballero, A.~Cunningham, A.~Acosta,
  A.~Aitken, A.~Tejani, J.~Totz, Z.~Wang \emph{et~al.}, ``Photo-realistic
  single image super-resolution using a generative adversarial network,'' in
  \emph{IEEE Conference on Computer Vision and Pattern Recognition}, 2017, pp.
  4681--4690.

\bibitem{yao2015color}
Q.~Yao and J.~T. Kwok, ``Colorization by patch-based local low-rank matrix
  completion,'' in \emph{Proceedings of the 29th {AAAI} Conference on
  Artificial Intelligence}, 2015, pp. 1959--1965.

\bibitem{zhang2016colorful}
R.~Zhang, P.~Isola, and A.~A. Efros, ``Colorful image colorization,'' in
  \emph{European Conference on Computer Vision}, 2016, pp. 649--666.

\bibitem{yeh2016semantic}
R.~Yeh, C.~Chen, T.~Y. Lim, M.~Hasegawa-Johnson, and M.~N. Do, ``Semantic image
  inpainting with perceptual and contextual losses,'' Preprint
  arXiv:1607.07539, 2016.

\bibitem{gatys2016image}
L.~A. Gatys, A.~S. Ecker, and M.~Bethge, ``Image style transfer using
  convolutional neural networks,'' in \emph{IEEE Conference on Computer Vision
  and Pattern Recognition}, 2016, pp. 2414--2423.

\bibitem{li2017universal}
Y.~Li, C.~Fang, J.~Yang, Z.~Wang, X.~Lu, and M.-H. Yang, ``Universal style
  transfer via feature transforms,'' in \emph{Advances in Neural Information
  Processing Systems}, 2017, pp. 385--395.

\bibitem{johnson2016perceptual}
J.~Johnson, A.~Alahi, and L.~Fei-Fei, ``Perceptual losses for real-time style
  transfer and super-resolution,'' in \emph{European Conference on Computer
  Vision}, 2016, pp. 694--711.

\bibitem{goodfellow2014generative}
I.~Goodfellow, J.~Pouget-Abadie, M.~Mirza, B.~Xu, D.~Warde-Farley, S.~Ozair,
  A.~Courville, and Y.~Bengio, ``Generative adversarial nets,'' in
  \emph{Advances in Neural Information Processing Systems}, 2014, pp.
  2672--2680.

\bibitem{gulrajani2017improved}
I.~Gulrajani, F.~Ahmed, M.~Arjovsky, V.~Dumoulin, and A.~C. Courville,
  ``Improved training of wasserstein gans,'' in \emph{Advances in Neural
  Information Processing Systems}, 2017, pp. 5769--5779.

\bibitem{mirza2014conditional}
M.~Mirza and S.~Osindero, ``Conditional generative adversarial nets,'' Preprint
  arXiv:1411.1784, 2014.

\bibitem{salimans2016improved}
T.~Salimans, I.~Goodfellow, W.~Zaremba, V.~Cheung, A.~Radford, and X.~Chen,
  ``Improved techniques for training gans,'' in \emph{Advances in Neural
  Information Processing Systems}, 2016, pp. 2234--2242.

\bibitem{arjovsky2017wasserstein}
M.~Arjovsky, S.~Chintala, and L.~Bottou, ``Wasserstein generative adversarial
  networks,'' in \emph{Proceedings of the 34th International Conference on
  Machine Learning}, 2017, pp. 214--223.

\bibitem{liu2017unsupervised}
M.-Y. Liu, T.~Breuel, and J.~Kautz, ``Unsupervised image-to-image translation
  networks,'' in \emph{Advances in Neural Information Processing Systems},
  2017, pp. 700--708.

\bibitem{zhu2017toward}
J.-Y. Zhu, R.~Zhang, D.~Pathak, T.~Darrell, A.~A. Efros, O.~Wang, and
  E.~Shechtman, ``Toward multimodal image-to-image translation,'' in
  \emph{Advances in Neural Information Processing Systems}, 2017, pp. 465--476.

\bibitem{kingma2014adam}
D.~Kingma and J.~Ba, ``Adam: A method for stochastic optimization,'' Preprint
  arXiv:1412.6980, 2014.

\bibitem{liu2015deep}
Z.~Liu, P.~Luo, X.~Wang, and X.~Tang, ``Deep learning face attributes in the
  wild,'' in \emph{IEEE International Conference on Computer Vision}, 2015, pp.
  3730--3738.

\bibitem{ronneberger2015u}
O.~Ronneberger, P.~Fischer, and T.~Brox, ``U-net: Convolutional networks for
  biomedical image segmentation,'' in \emph{International Conference on Medical
  Image Computing and Computer-Assisted Intervention}, 2015, pp. 234--241.

\bibitem{he2016deep}
K.~He, X.~Zhang, S.~Ren, and J.~Sun, ``Deep residual learning for image
  recognition,'' in \emph{IEEE Conference on Computer Vision and Pattern
  Recognition}, 2016, pp. 770--778.

\bibitem{li2016precomputed}
C.~Li and M.~Wand, ``Precomputed real-time texture synthesis with markovian
  generative adversarial networks,'' in \emph{European Conference on Computer
  Vision}, 2016, pp. 702--716.

\bibitem{kingma2013auto}
D.~P. Kingma and M.~Welling, ``Auto-encoding variational bayes,'' Preprint
  arXiv:1312.6114, 2013.

\bibitem{kingma2016improved}
D.~P. Kingma, T.~Salimans, R.~Jozefowicz, X.~Chen, I.~Sutskever, and
  M.~Welling, ``Improved variational inference with inverse autoregressive
  flow,'' in \emph{Advances in Neural Information Processing Systems}, 2016,
  pp. 4743--4751.

\bibitem{maaloe2016auxiliary}
L.~Maal{\o}e, C.~K. S{\o}nderby, S.~K. S{\o}nderby, and O.~Winther, ``Auxiliary
  deep generative models,'' in \emph{Proceedings of the 33rd International
  Conference on Machine Learning}, 2016, pp. 1445--1453.

\end{thebibliography}

\end{document}